\documentclass[times,twocolumn,final,authoryear]{elsarticle}

\usepackage{ycviu}
\usepackage{framed,multirow}

\usepackage{amssymb}
\usepackage{latexsym}

\usepackage{url}
\usepackage{xcolor}

\usepackage{algorithmic}
\usepackage{algorithm}

\usepackage{multirow}
\usepackage{multicol}
\usepackage{makecell}

\definecolor{newcolor}{rgb}{.8,.349,.1}

\journal{Computer Vision and Image Understanding}

\begin{document}

\begin{frontmatter}

\title{Hybrid 3D Human Pose Estimation with Monocular Video and Sparse IMUs}

\author[1]{Yiming \snm{Bao}} 
\author[2]{Xu \snm{Zhao}\corref{cor1}}
\cortext[cor1]{Co-corresponding authors.}
\ead{zhaoxu@sjtu.edu.cn;dahong.qian@sjtu.edu.cn}
\author[1]{Dahong \snm{Qian}\corref{cor1}}

\address[1]{the School of Biomedical Engineering, Shanghai Jiao Tong University, Shanghai, China}
\address[2]{the Department of Automation, Shanghai Jiao Tong University, Shanghai, China}

\received{1 May 2013}
\finalform{10 May 2013}
\accepted{13 May 2013}
\availableonline{15 May 2013}
\communicated{S. Sarkar}

\begin{abstract}
Temporal 3D human pose estimation from monocular videos is a challenging task in human-centered computer vision due to the depth ambiguity of 2D-to-3D lifting. To improve accuracy and address occlusion issues, inertial sensor has been introduced to provide complementary source of information. However, it remains challenging to integrate heterogeneous sensor data for producing physically rational 3D human poses. In this paper, we propose a novel framework, Real-time Optimization and Fusion (RTOF), to address this issue. We first incorporate sparse inertial orientations into a parametric human skeleton to refine 3D poses in kinematics. The poses are then optimized by energy functions built on both visual and inertial observations to reduce the temporal jitters. Our framework outputs smooth and biomechanically plausible human motion. Comprehensive experiments with ablation studies demonstrate its rationality and efficiency. On Total Capture dataset, the pose estimation error is significantly decreased compared to the baseline method.
\end{abstract}


\end{frontmatter}

\author{
Yiming Bao, Xu Zhao*, \IEEEmembership{Member, IEEE}, Dahong Qian*, \IEEEmembership{Senior Member, IEEE}
\thanks{X. Zhao and D. Qian are co-corresponding authors.}
\thanks{Y. Bao and D. Qian are with  (e-mail: yiming.bao@sjtu.edu.cn; dahong.qian@sjtu.edu.cn)}
\thanks{X. Zhao is with  (e-mail: )}
\thanks{This work has been supported by the NSFC grants 62176156 and Deepwise Healthcare Joint Research Lab, Shanghai Jiao Tong University.}
}

\section{Introduction}
\label{sec:intro}
Monocular 3D human pose estimation (HPE) had been wildly studied for many years. It can be applied for many downstream scenes (\cite{gao2019dual-hand,guo2022self}). Researchers follow a two-step way to reconstruct 3D human pose from video, i.e., first detecting 2D keypoints in images and then lifting them to 3D space. Recently, many approaches propose to train more robust 2D keypoints detectors (\cite{xiao2018simple,sun2019deep}) to reduce the 2D estimation error. In contrast, the 3D reconstruction error is relatively more difficult to reduce. Some methods attempt to triangulate multi-view 2D information (\cite{iskakov2019learnable,qiu2019cross}). However, multiple-camera is restricted to indoor settings thus limiting the real-scene applications.

Recently, some methods (\cite{videopose3d,chen2021anatomy,poseformer,shan2022p}) take 2D pose sequences as input and train a many-to-one frame aggregator to predict 3D pose in the center frame. These aggregators can learn spatial and temporal correlations from the 2D pose sequences. Despite the achieved considerable development, 3D HPE from monocular videos remains challenging since the 2D-to-3D lifting is inherently ill-conditioned due to the depth ambiguity and occlusions. Moreover, the regressed 3D pose sequences still suffer from artifacts such as jitters and skating. 

To solve the above-mentioned issues, inertial sensors such as Inertial Measurement Unit (IMU) are valuable as they can provide high-frequency while accurate and occlusion-free 3D local information (\cite{pons2014human,von2016human}). To reach this, the key is to efficiently integrate heterogeneous visual and inertial data.

In this work, we propose a framework called Real-Time Optimization and Fusion (RTOF), to fuse monocular video and IMUs spatially and temporally for physically more plausible 3D pose sequences. Under a hybrid sensor fusion paradigm, RTOF is able to 1) alleviate the influence of occlusions and depth ambiguity; 2) greatly reduce the local error and jitter after 2D-to-3D lifting; 3) serve as a valuable solution in outdoor and real-time applications. Specifically, it first takes 2D pose sequences as input and predicts 3D pose sequences by learning 2D correlations across consecutive frames. Then, inertial data is exploited under a parametric human model to refine the lifted 3D pose in kinematic space. Different from previous approaches which utilize transformed IMU data (\cite{huang2020deepfuse,zhang2020fusing}), our strategy directly aggregates the raw inertial containing accurate local information. Finally, to further reduce the temporal jitter, a hybrid and comprehensive energy function is optimized to fit the observation of both 2D poses and inertial measurements. Existing work (\cite{von2018recovering}) optimizes the whole sequence which is time-consuming. Hence, we propose a fragment-based strategy to match the real-time requirements.

Experiments demonstrate that RTOF can estimate 3D pose sequences with high spatial accuracy and temporal smoothness. On Total Capture dataset, our method reduces the MPJPE from $64.6mm$ of the initial lifted 3D pose sequence result to $33.7mm$, benefiting from the sensor fusion and temporal optimization. When fed with monocular 2D pose sequences with no 2D error for lifting, our method gets a result of $23.2mm$. This even outperforms the state-of-the-arts with multi-view input, which indicates the great potential of our method with monocular video and sparse IMUs input. On Human3.6M dataset, our visual-only strategy also achieves comparable results compared to the previous methods, especially on the metrics for temporal error. To sum up, the main contributions provided in this paper are as follows:

\indent $\bullet$ We propose a hybrid framework that performs 3D HPE with monocular video and sparse IMUs and matches the real-time requirements.

\indent $\bullet$ We design a strategy enabling efficient visual-inertial fusion in the human kinematic space as well as at the temporal level, resulting in physically plausible human poses.

\indent $\bullet$ The proposed framework significantly improves the performance of 3D HPE temporally and spatially, reaching a comparable level with multi-view methods.

\section{Related Work}
\subsection{Temporal 3D HPE}
The main purpose of 3D HPE is typically to regress the 3D human joint locations (\cite{sun2018integral}) or rotations (\cite{loper2015smpl,li2021hybrik,shi2020motionet,bao2022fusepose}) of a predefined skeleton. A majority of approaches first regress 2D joint locations in the image plane and subsequently lift them to the camera or world 3D space (\cite{zhan2022ray3d,martinez2017simple}). These two-step methods substantially outperform the end-to-end counterparts that directly regress 3D human pose from the image.

The early method (\cite{martinez2017simple}) proposes to directly perform 2D-to-3D lifting in a single-frame setting, which offers a baseline and paradigm for 2D-to-3D lifting. Recently, more studies exploit the value of temporal correlation within the 2D sequence to improve the robustness. \cite{hossain2018exploiting} proposed an RNN to learn from the temporal input. \cite{videopose3d} introduced a pioneering temporal convolution network (TCN) to estimate 3D pose of the center frame. Based on \cite{videopose3d}, \cite{chen2021anatomy} further designed a bone direction module and a bone length module to maintain the temporal consistency of human anatomy. Other methods (\cite{poseformer,shan2022p}) exploit the transformer-based architectures to model the temporal and spatial correlations.
      
The above-mentioned 2D-to-3D lifting methods aggregate the information in the input 2D pose sequence while the lifted 3D pose sequence result has not been exploited. In this work, we prove it valuable to optimize the lifted 3D pose sequence in an online manner.

\subsection{Generative Pose Optimization}
Before learning-based methods took the main place, the generative models were popular and dominant in 3D HPE (\cite{lassner2017unite,liu2013markerless,helten2013personalization}). The generative model-based algorithms aim at optimizing the energy errors by fitting 3D human pose or parameterized human skeleton to the observations. The required observations are mostly from images and image features, sometimes also from other sensors such as IMUs. The human skeleton can be parameterized as Euler angle (\cite{eulerangle}), Exponential Map (\cite{exponentialmap}) or Quaternion (\cite{quaternion,pavllo2019modeling}). This parameterization of 3D human pose enables reliable initialization such as T-pose for better fitting the predicted results. In other words, the search space can be simplified and the search efficiency can be facilitated. 

Generally, the majority of this category of algorithms perform the optimization process on single frame (\cite{single1,single2,single3}) or on multiple frames of the motion (\cite{multiframe1,multiframe2,multiframe3,multiframe4}). However, single frame optimization brings jitters and also heavily relies on the initialization while multiple frames optimization is time-consuming and thus inefficient for the real-time scene. In this work, we demonstrated an efficient optimization strategy to predict fragments of human poses in an online manner.

\begin{figure*}[!ht]
 \begin{center}
 	\centerline{\includegraphics[width= \linewidth]{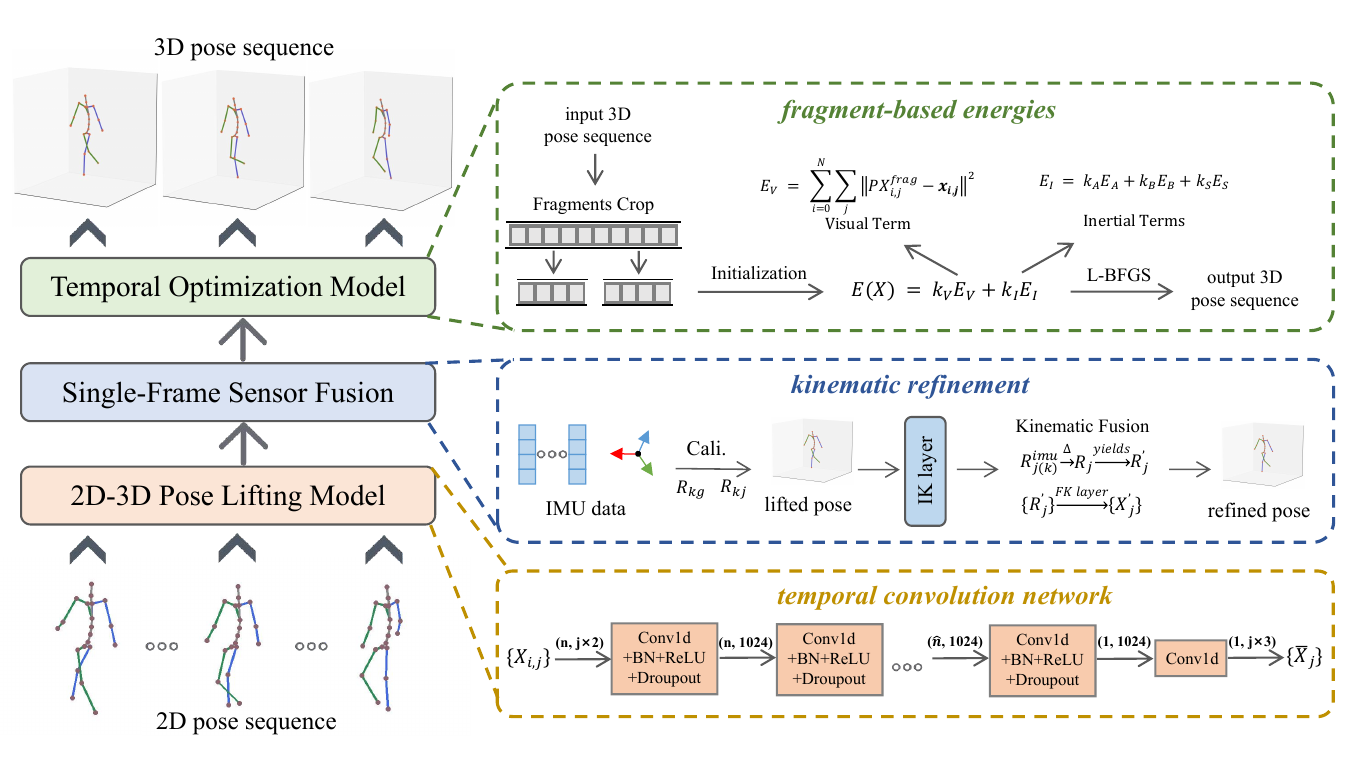}}
\caption{Overview of the proposed RTOF framework. Three modules are successively inferred. First, 2D pose sequence is lifted to 3D space with a receptive field of $n$ frames. The single frame sensor fusion is then conducted to refine the lifted 3D pose using calibrated and aligned IMU raw data. Finally, fragments with $N$ frames are cropped from the refined 3D sequence and optimized by both visual and inertial energy functions.}
\label{fig_overview}
\end{center}
\end{figure*}

\subsection{Vision-Inertial Fusion for 3D HPE}
Inertial sensors are free of the influence caused by occlusion or light changing when performing 3D HPE. Thus, they are wildly exploited for improving the performance (\cite{von2016human,huang2020deepfuse,pons2010multisensor,zhang2020fusing}). Inertial sensors measure motion accelerations and angular velocities. The orientations or other data are further calculated and optimized via filter algorithms (\cite{vitali2020robust}). The commercial solution of the Inertial Measurement Unit (IMU) (\cite{xsens2009full}) can be conveniently mounted on human bones for 3D HPE. 

Albeit inertial data is valuable to aggregate for correcting bone direction or joint accuracy, it is challenging to fuse it with the visual intermediate features such as lifted 2D or 3D pose. The existing algorithms tend to transform the IMU orientation to bone vectors (\cite{zhang2020fusing}), or just perform the fusion in a learning-based neural network (\cite{huang2020deepfuse}). These methods ignore the raw IMU data so they can hardly achieve well-refined 3D pose or motion and the IMU acceleration data has not been sufficiently employed. Moreover, most of the previous works (\cite{zhang2020fusing,bao2022fusepose}) explore how IMU can refine the multi-view 3D pose estimation results while not the monocular output. Despite the high performance achieved with the multi-view input, these methods are not applicable to outdoor and other common scenes. Thus, in this work, we refine the lifted 3D pose from single view using the raw IMU data, i.e., both the IMU orientation and acceleration, in a tight and reasonable manner.


\section{Method}
\label{sec:method}

\subsection{Overview}
\label{subsec:overview}
In this section, we introduce the proposed framework, which is consisted of three main parts: 2D-to-3D lifting model, single frame sensor fusion ($SF^{2}$), and temporal optimization model. The overview of the proposed method is illustrated in Figure \ref{fig_overview}. There are two main ways to employ our framework. The first one is inferring all three parts of it with both visual and inertial sequence, referred to as real-time optimization and fusion (RTOF). The second way is only utilizing the visual clue while skipping the second $SF^{2}$ part and also not using inertial information in the third part of the framework, referred to as real-time optimization (RTO).

\subsection{2D-to-3D Pose Lifting Model}

The IMU data cannot be directly fused with the 2D human pose or image frames. The mapping between the measured inertial rotations and the 2D imaging data is very difficult to model or learn. An appropriate solution is to lift the 2D pose to 3D space and then fuse visual and inertial features.

In this work, we adopt a temporal convolution network proposed in (\cite{videopose3d}) as the backbone architecture of our 2D-to-3D lifting model. Specifically, the 2D keypoints of consecutive frames are concatenated to form the input of the network. The network architecture is illustrated in Figure \ref{fig_overview} and more details are referenced to (\cite{videopose3d}). The 2D joint coordinates sequence with $n$ frames is sampled from a 2D pose sequence, where for each frame $x_{i} \in R^{J \times 2}$. $J$ is the number of joints. The model predicts 3D pose for the center frame $X \in R^{J \times 3}$. Note that the 2D pose is directly lifted to the global coordinate. The 2D-to-3D lifting model is trained using the joint error loss which minimizes the difference between the predicted 3D pose and the ground truth.

The 3D pose predicted by the 2D-to-3D lifting model is inaccurate in the depth direction due to ambiguity in the back-projection process. IMU data can be utilized for 3D pose refinement. However, it is still hard to directly fuse the two different modalities of data explicitly or implicitly. In this work, we propose to transfer the lifted 3D pose to human kinematic space to align and fuse visual and inertial information explicitly.

\subsection{Single Frame Sensor Fusion}

Previous methods (\cite{zhang2020fusing}) transform IMU information to bone vectors and then utilize them for a more accurate 3D pose. However, this strategy ignores the raw data of IMU for deeper sensor fusion. In this work, we design a tightly entangled method to perform single-frame sensor fusion ($SF^{2}$) in human kinematic space. Our method not only allows the use of IMU raw data but also brings human skeleton consistency in the 3D pose sequence.

First, we define a 3D human skeleton with local motion parameters. Given the root joint transition $\mathcal{T}_{0}$, the rotation $\mathcal{R} = \lbrace R_{j} \rbrace_{j=1}^{J}$ of each joint local coordinate system relative to the rest T-pose, the 3D human pose can be represented with either the 3D location of all joints $\mathcal{X} = \lbrace X_{j} \rbrace_{j=0}^{J}$ or the motion parameters $\lbrack \mathcal{T}_{0}, \mathcal{R}\rbrack$. These two representations can be transformed with each other using forward kinematics and inverse kinematics \cite{csiszar2017solving}. For the purpose of integrating the IMU information into the parametric representation of the 3D pose, the IMU-recorded rotation data needs to be aligned to the local human joint coordinates. We first employ the global joint rotation $R_{j(k)}^{global}$ of the $j$th joint which is mounted by the IMU $k$ on its controlled bone, the recorded data $R_{k}$ of the IMU needs to be calibrated first by the IMU reference frame-global offset $R_{kg}$, and then by the offset $R_{kj}$ between the IMU reference and the coordinate system of the corresponding joint $j$. The IMU-measured global joint rotation $R_{j(k)}^{imu}$ for global joint rotation can be calculated by:
\begin{equation}
R_{j(k)}^{imu} = (R_{kj})^{-1} R_{kg} R_{k},
\end{equation}
where $R_{j(k)}^{imu}$ can be further utilized in the proposed Inertial-Guided Inverse Kinematic (IGIK) layer to refine the solved global joint rotation $R_{j(k)}^{global}$. Finally, we obtain the predicted motion rotation parameters $\mathcal{R}$ that matches both the visual and inertial observation. The detail of IGIK layer is illustrated in the Alg. \ref{alg1}.

\begin{algorithm}[t] 
\caption{IGIK layer}
\begin{algorithmic}
    \STATE \textbf{Input:} $\mathcal{X}, R^{imu}, \mathcal{X}^{T}$
\STATE \textbf{Output:} $\mathcal{R}$
\STATE \hspace{0.5cm} \textbf{for} $j$ along the kinematic tree \textbf{do}
\STATE \hspace{1cm} $B_{j} \gets (X_{j} - X_{pa(j)})$
\STATE \hspace{1cm} $R_{j}^{global} \gets$ SolveRotation($B_{j}^{T}, B_{j}$)
\STATE \hspace{1cm} \textbf{if} joint $j$ is attached with IMU $k$ \textbf{then}
\STATE \hspace{1.5cm} $B_{k}^{imu} \gets R_{j(k)}^{imu} \rhd B_{j}$
\STATE \hspace{1.5cm} $\theta_{k} \gets$  $<B_{j(k)}^{imu}, B_{j}>$
\STATE \hspace{1.5cm} \textbf{if} $\theta_{k} > \theta_{t}$ \textbf{then}
\STATE \hspace{2.0cm} $R_{j}^{global} \gets R_{j(k)}^{imu}$
\STATE \hspace{1.5cm} \textbf{end if}
\STATE \hspace{1cm} \textbf{end if}
\STATE \hspace{1cm} $R_{j}^{local} \gets (R_{pa(j)}^{global})^{-1} \otimes R_{j}^{global}$
\STATE \hspace{0.5cm} \textbf{end for}
\STATE \hspace{0.5cm} $R \gets \lbrace R_{j}^{local} \rbrace_{j=1}^{J}$
\end{algorithmic}
\label{alg1}
\end{algorithm}

The calibrated IMU rotation is aggregated to revise the local joint rotation, which ensures that the human skeleton is controlled not only by the prediction from the visual clue but also by the inertial measurement. Moreover, the improvement of local rotation estimation for upstream joints in the human skeleton tree can further promote the estimation for lower downstream joints. This is plausible because the human skeleton is a cascading chain with anatomical constraints. After all the used IMUs are employed for human motion refinement, the human pose is transformed back to 3D location representation via forward kinematics. Besides the revision of local rotation estimation, another noteworthy value of this kinematic sensor fusion is that, the consistency of the human skeleton can be maintained for all frames. To be specific, the bone lengths remain unchanged for all 3D poses of the same motion. This solves the bone stretching issue and also brings estimation error improvement for those joints not mounted with IMUs.

\begin{figure}[t]
 \begin{center}
 	\centerline{\includegraphics[width= 6cm ]{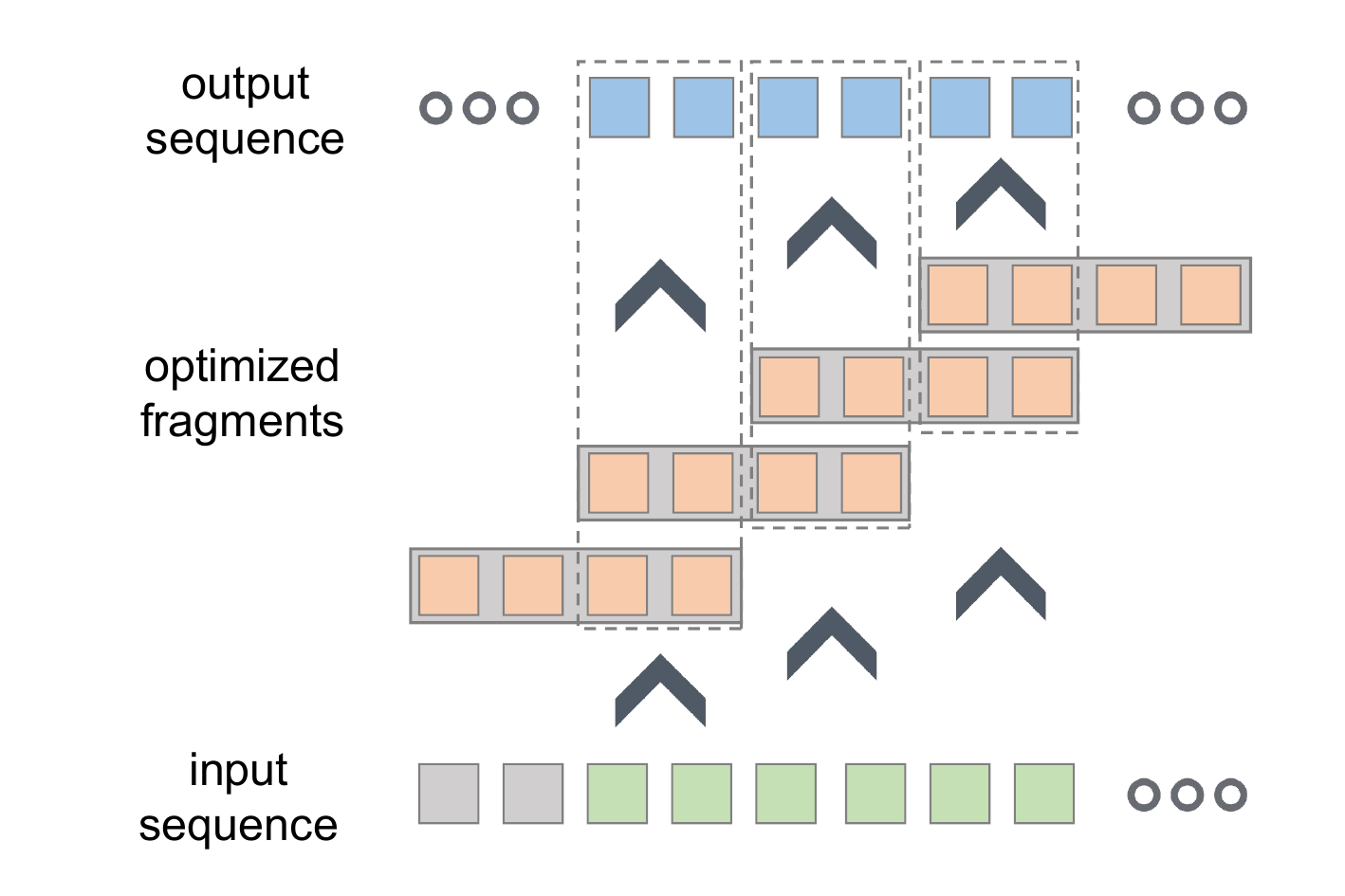}}
\caption{An illustration of the proposed fragment cropping approach with $N=4$. In this case, the crop step and the length for the final average are both $2$. The cropped fragments are optimized by the temporal optimization model and are then utilized to calculate the output 3D pose sequence.}
\label{fig_fragment}
\end{center}
\end{figure}

\begin{table*}[!ht]
\caption{Comparison of the 3D pose estimation errors MPJPE (mm) of different methods on the Total Capture dataset. RTOF-SN outperforms the baseline VideoPose3D by a large margin. RTOF-GT even surpasses the multi-view input methods.}
\begin{center}
\resizebox{\textwidth}{!}{
\begin{tabular}{lcccccccccc}
\hline
Method & Cameras & Train & Test&\multicolumn{3}{c}{SeenSubject (S1,2,3)} &\multicolumn{3}{c}{UnseenSubject (S4,5)} & Average  \\
 &&w/ IMUs&w/ IMUs&W2&A3&FS3&W2&A3&FS3&\\
\hline
LSTM-AE \cite{trumble2018deep} &4 & & &13.0 &23.0&47.0 &21.8 &40.9 &68.5 &34.1\\
IMUPVH \cite{gilbert2019fusing} &8 &\checkmark&\checkmark&19.2 &42.3&48.8 &24.7 &58.8 &61.8 &42.6\\
Fusion-RPSM \cite{qiu2019cross} &4 & & &19.0 &21.0&28.0 &32.0 &33.0 &54.0 &29.0\\
LWCDR \cite{remelli2020lightweight} &4 & & &\textbf{10.6} &16.3&\textbf{30.4} &27.0 &\textbf{34.2} &65.0 &27.5\\
DeepFuse \cite{huang2020deepfuse} &8 &\checkmark &\checkmark&- &-&- &- &- &- &28.9\\
GeoFuse \cite{zhang2020fusing} & 4 &\checkmark&\checkmark&14.3 &17.5 & 25.9 &23.9 &27.8 &49.3 &24.6\\
VIP \cite{von2018recovering} & 1 &\checkmark&\checkmark&- &- & - &- &- &- &26.0\\
RTOF-GT (Ours) & 1 & & \checkmark &10.8&\textbf{13.9}&32.6&\textbf{16.9}&34.8&\textbf{41.5}&\textbf{23.2}\\
\hline
VP3D-SN \cite{videopose3d} & 1 & & &23.6&43.6&89.9&49.1&87.8&123.6&64.6\\
RTOF-SN (Ours) & 1 & & \checkmark &15.7&20.1&47.3&24.5&50.5&60.2&33.7\\
\hline
\end{tabular}}
\end{center}
\label{table_tc_sota}
\end{table*}

\subsection{Temporal Optimization Model}

Optimization models for human poses have been popular for decades since they are extensible and effective when observations are reliable. In this work, we first consider the 2D pose sequence as the indispensable observation. Additionally, we include IMU data for establishing the final energy function and optimizing the 3D pose sequence in temporal. We also propose a fragment-based strategy, which enables online operation and output 3D pose sequences in real-time.

The parameters to optimize are the 3D joint locations of a fragment with $N$ consecutive frames $X^{frag} \in R^{N \times J \times 3}$. $X^{frag}$ is cropped from the outputted 3D pose sequence of $SF^{2}$ module. To avoid the discontinuity caused by slicing, we propose a novel fragment cropping approach as illustrated in Figure. \ref{fig_fragment} which shows an example of fragment crop and optimization when $N=4$. Starting from the beginning of the input sequence where we pad with replicas of the boundary frames, the cropping is performed with sliding step as $\frac{N}{2}$. Thus, each frame of the input sequence can be optimized twice and then averaged in the final output sequence, which avoids the potential discontinuity in all frames.

Based on the cropped fragment, we build a comprehensive and hybrid energy function to optimize the errors between observation and prediction. The function consists of the visual term and the inertial term as follows:
\begin{equation}
E(X^{frag}) = k_{V}E_{V} + k_{I}E_{I},
\end{equation}where $k_{V}$ and $k_{I}$ are the weight of the visual energy term and the inertial energy term, respectively. The visual energy is built as the re-projection error as follows:
\begin{equation}
E_{V} = \sum_{i=0}^{N} \sum_{j=0}^{J} || P X_{i,j}^{frag} - x_{i,j} ||^{2},
\end{equation}where $P$ is the projection matrix of the camera. $x_{i,j}$ is the 2D joint location of the $j$-th joint in $i$-th frame and $X_{i,j}^{frag}$ is the 3D joint location in a fragment to optimize. This energy can partially reduce the estimation error. However, the impact caused by the depth ambiguity in 2D-to-3D lifting still exists. Thus, we propose to further build the inertial energy using IMU acceleration and orientation. The inertial energy term is built as follows:
\begin{equation}
E_{I} = k_{A}E_{A} + k_{B}E_{B} +  k_{S}E_{S},
\end{equation}where $E_{A}$, $E_{B}$ and $E_{S}$ are acceleration term, bone vector term and smooth term, respectively. $k_{A}$,$k_{B}$ and $k_{S}$ are their weights. Specifically, each of them is built as follows:
\begin{equation}
E_{A} = \sum_{i=0}^{N} \sum_{k} || A_{i,j(k)}^{frag} - A_{i,k}^{imu} ||^{2},
\end{equation}where $A_{i,j(k)}^{frag}$ is the second-order differential of the predicted 3D pose fragment for the joint $j$ mounted with the IMU $k$ in the frame $i$. The observed acceleration $A_{i,k}^{imu}$ comes from the recorded IMU data $A_{i,k}^{rec}$ with calibration as follows:
\begin{equation}
A_{i,k}^{imu} = R_{kg}R_{i,k}A_{i,k}^{rec} - g,
\end{equation}where $R_{kg}$ is the transformation from IMU local to global and $R_{i,k}$ is the IMU recorded rotation w.r.t its reference frame. Similarly, the IMU orientation is switched to bone vectors and utilized to build a bone vector term as follows:
\begin{equation}
E_{B} = \sum_{i=0}^{N} \sum_{k} || B_{i,j(k)}^{frag} - B_{i,k}^{imu} ||^{2},
\end{equation}where $B_{i,k}^{imu}=R_{i,k}B^{T}_{k}$ and $B_{k}^{T}$ is from the rest T-pose, while the predicted bone vector are calculated from the 3D poses via $B_{i,j(k)}^{frag} = X^{frag}_{i,j(k)} - X^{frag}_{i,pa(j)}$ and $pa(j)$ is the parent joint of joint $j$ in the predefined human skeleton. Finally, for reducing the jitters in the output 3D pose sequence, we build a smooth term by:
\begin{equation}
E_{S} = \sum_{i=0}^{N} \sum_{k} || S_{i,j(k)}^{frag} - S_{i,k}^{imu} ||^{2},
\end{equation}where $S^{frag}$ and $S^{imu}$ are the first-order differentials of $A^{frag}$ and $A^{imu}$, respectively.

The optimization of the comprehensive energy function is implemented in PyTorch (\cite{paszke2019pytorch}) using L-BFGS (\cite{liu1989limited}) and \emph{autograd} (\cite{paszke2017automatic}).


\section{Experiments}

\subsection{Datasets and Evaluation Metrics}
\noindent \textbf{Total Capture} (\cite{trumble2017total}).
\quad This dataset is a large-scale benchmark containing information of 13 IMUs, videos and 3D human pose ground truth. We partition the training and testing dataset according to subjects and performance sequence. The training set is used for training the 2D-to-3D lifting model. It consists of performances ROM1, 2, 3; Walking1, 3; Freestyle1, 2; Acting1, 2 and Running1 on subjects 1, 2 and 3. The testing set contains the performances Freestyle3, Acting3 and Walking2 on all subjects. For 3D poses, we adopt 21 joints for the skeleton consistency of the subject in motion. We train a single 2D-to-3D lifting model for all actions and subjects. The IMU information is calibrated using the methods described in Sec \ref{sec:method}.

\noindent \textbf{Human3.6M} (\cite{ionescu2013human3}).
\quad This dataset consists of 3.6M frames of videos of 11 subjects. We split the training and testing set and adopt 17 joints following previous works (\cite{sun2018integral,zhang2020fusing}). Since it has no IMU sensors, so we just utilize this dataset for evaluating the fragment-based optimization model with only visual energy, i.e., RTO.

\noindent \textbf{Evaluation Metrics.}
\quad We first consider the mean per joint position error (MPJPE) in millimeters. Second, we consider the mean per joint acceleration error (MPJAE) which evaluates the accuracy of the second-order differential of the predicted 3D pose per second. Finally, to evaluate the jitters in the output 3D pose sequence, we calculated the mean per joint jitter error (MPJJE) from the first-order differential of joint acceleration. MPJJE can directly demonstrate whether the predicted motion is smooth or not.

\noindent \textbf{Implementation details.}
\quad The training strategy of the 2D-to3D lifting model mainly follows \cite{videopose3d} which randomly samples chunks with a receptive field of 27 frames. We adopt Adam optimizer (\cite{kingma2014adam}) with a learning rate of $0.05$ and train the model for $1500$ epochs. For sensor fusion, we use 8 sparse IMUs mounted on limbs. When testing, the 2D joint locations are acquired from a pre-trained 2D pose estimation backbone from \cite{xiao2018simple}, referred to as SimpleNet (SN). In the temporal optimization stage, we first normalize each energy term in the same order of magnitude to balance their contribution to the optimization process. Then, the weights of different sensor terms are empirically set as $k_{V} = 0.9$, $k_{I} = 0.1$. For the inertial terms, we set them as $k_{A} = 0.5$, $k_{B} = 0.2$, $k_{J} = 0.3$. The length of the cropped fragments is set as $N=50$.

\subsection{Comparison to Other Methods}
\noindent \textbf{RTOF on Total Capture.}
\quad We first compare our proposed RTOF framework with the state-of-the-art methods on Total Capture dataset. As shown in Table \ref{table_tc_sota}, our RTOF achieves remarkable performance. First, compared to the baseline result of VP3D-SN which also inputs SimpleNet estimated 2D pose to the baseline model VP3D (\cite{videopose3d}), RTOF-SN decreases the estimation error from $64.6mm$ to $33.7mm$. This large margin of improvement is brought by both the sensor fusion and the temporal optimization. The respective contribution of each of them will be analyzed in the ablation study part.

We can also note that most previous methods estimate 3D poses from multi-view input. Multi-view complementations for reconstruction decrease the effect of depth ambiguity to a negligible extent with triangulation (\cite{iskakov2019learnable}). Thus, it is unfair to directly compare RTOF-SN with them. Following previous monocular input methods (\cite{gong2021poseaug,poseformer}), we report another result RTOF-GT using ground truth 2D pose sequence input. This result isolates the influence of 2D pose estimation error so it can purely evaluate the 2D-to-3D lifting model. Also, this narrows the input gap between previous multi-view methods with ours. It can be observed that RTOF-GT achieves the result of $23.2mm$ and outperforms all of the multi-view methods, which strongly proves the effectiveness and advantage of our RTOF framework.

\noindent \textbf{Compared to inertial-only methods.}
\quad We also compare our fusion strategy with some inertial-only methods which use more sparse IMUs for 3D human pose estimation. The results are summarized in Table \ref{table_imu}. Note that other methods use 6 IMUs and reconstruction SMPL poses from AMASS dataset \cite{mahmood2019amass}, which are different from the skeletons predicted by our methods. Our method receives lower MPJPE through sensor fusion than previous IMU-only methods. Meanwhile, MPJJE is also comparable due to the proposed temporal optimization strategy.

\begin{table}
\caption{Comparison to IMU-only methods on Total Capture Datasets.}
\begin{center}
\scalebox{0.4}{
\resizebox{\textwidth}{!}{
\begin{tabular}{lcc}
\hline
Method & MPJPE & MPJJE\\
\hline
DIP \cite{huang2018deep} &87.6 &159.1\\
TransPose \cite{yi2021transpose} &54.2&77.2\\
TIP \cite{jiang2022transformer} &52.7& \textbf{53.6}   \\
RTOF-SN (ours) &\textbf{33.7}& 63.8\\
\hline
\end{tabular}}}
\end{center}
\label{table_imu}
\end{table}

\begin{table}[!t]
\caption{The results of 3D HPE metrics on Human3.6M dataset. The temporal smoothness of the output sequence is well improved. Here MPJAE and MPJJE are calculated per frame.}
\begin{center}
\scalebox{0.45}{
\resizebox{\textwidth}{!}{
\begin{tabular}{cccc}
\hline
Metric & MPJPE & MPJAE &MPJJE\\
\hline
\cite{videopose3d} &46.8&5.85&47.74\\ 
\cite{chen2021anatomy}&44.1&4.65&37.25 \\
RTO-SN (ours)&46.1&2.03&11.25\\
\hline
\end{tabular}}}
\end{center}
\label{table_h36m_jitter}
\end{table}

\noindent \textbf{RTO on Human3.6M.}
\quad On Human3.6M, we compare the visual-only strategy RTO with other state-of-the-arts as listed in Table \ref{table_h36m_jitter}. Compared to the baseline result in \cite{videopose3d}, our RTO strategy improves the result from $46.8mm$ to $46.1mm$. It can also be observed that the MPJPE of RTO is higher than another method \cite{chen2021anatomy}. This is mainly because our 2D-to-3D lifting backbone is the original \cite{videopose3d} while the competitor adds two more modules to it. 

RTO is also able to decrease the motion jitters which is an important unsolved issue in the real-scene application. To evaluate this, we calculate and compare acceleration error and jitter error with the baseline model \cite{chen2021anatomy}. Note that metrics are calculated per frame here, following \cite{chen2021anatomy}. We can observe that the results are well improved. This demonstrates that the RTO strategy substantially decreases the motion jitters and ensures smooth and plausible output. 

\begin{figure}[!t]
 \begin{center}
 	\centerline{\includegraphics[width= \linewidth]{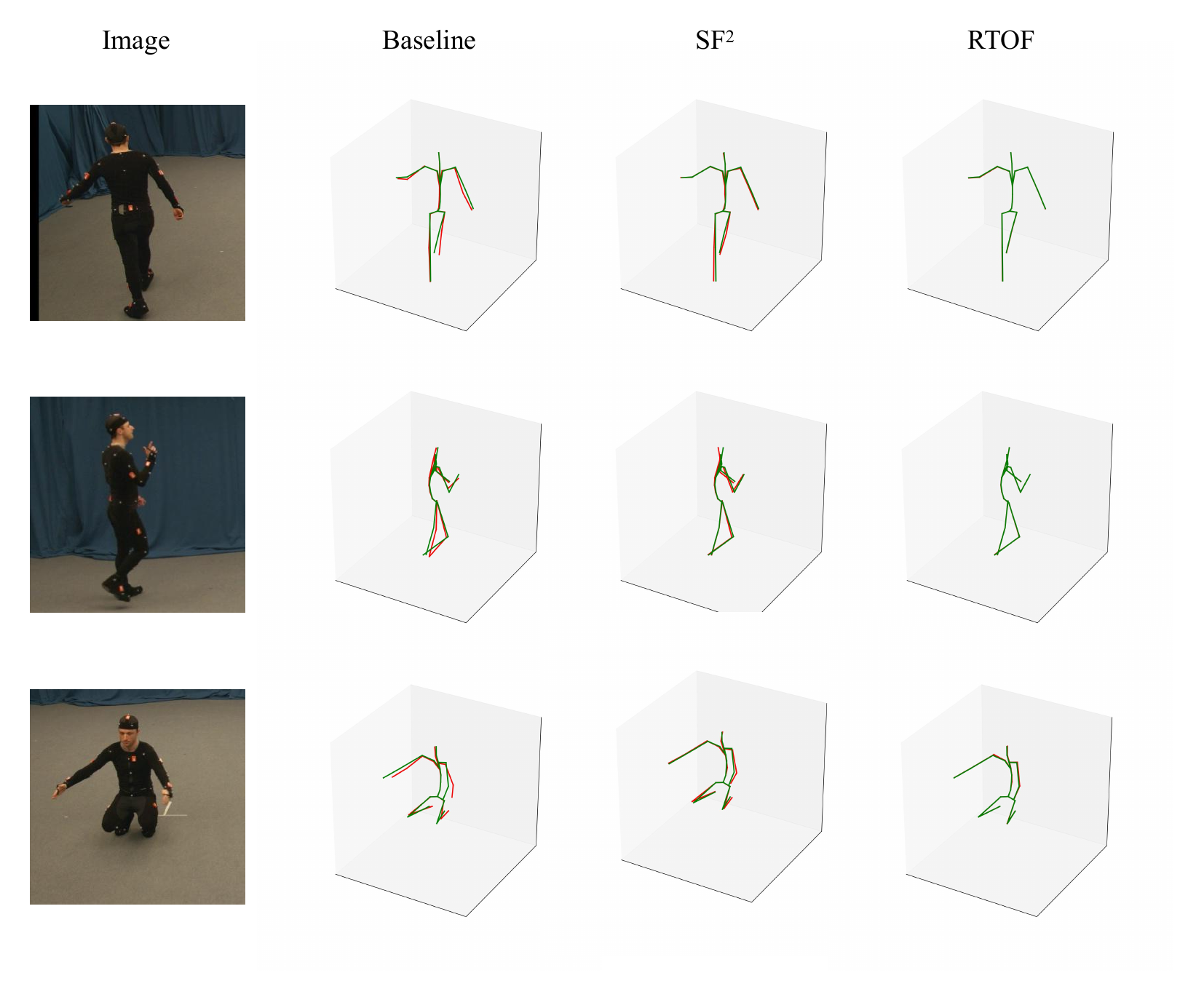}}
\caption{The illustration of the estimated 3D human poses (red skeletons) compared with ground truth (green skeletons).}
\label{fig_pose}
\end{center}
\vspace{-3em}
\end{figure}

\begin{figure*}[!ht]
 \begin{center}
 	\centerline{\includegraphics[width= \linewidth]{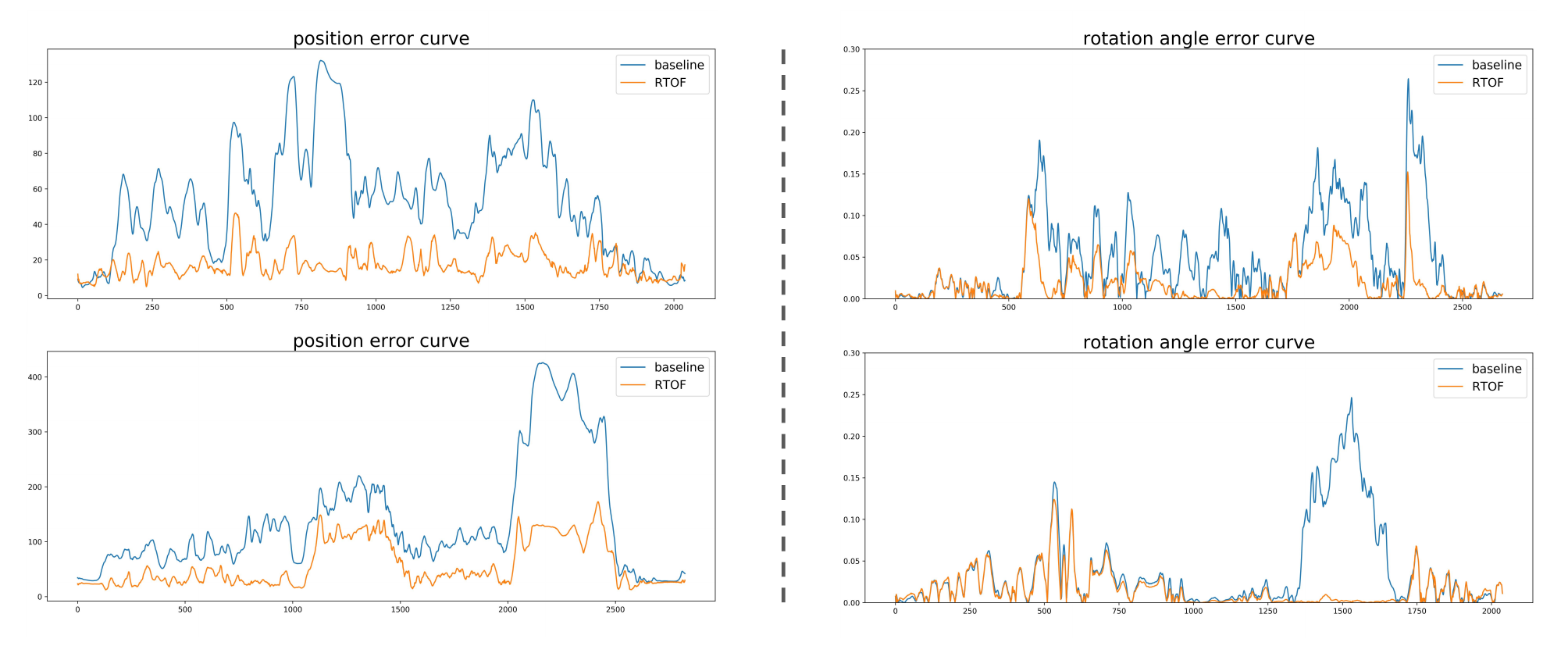}}
\caption{The quantitative ablation results on Total Capture dataset. The per joint position error curves and rotation angle error curves on all frames from four sequences with different motion characteristics are drawn.}
\label{fig_curves}
\end{center}
\end{figure*}

\subsection{Ablation Study}

\begin{table}[!t]
\caption{The quantitative ablation results of 3D HPE metrics on Total Capture dataset. Here MPJAE and MPJJE are calculated per second using ground truth 2D poses.}
\begin{center}
\scalebox{0.35}{
\resizebox{\textwidth}{!}{
\begin{tabular}{lccccc}
\hline
Metric & Baseline & RTO & $SF^{2}$ & RTOF\\
\hline
MPJPE &53.7 &40.6&34.5&23.2 \\
MPJAE &2.70&1.79 &3.96&1.54\\
MPJJE &87.3& 55.2&185.4 &63.8\\
\hline
\end{tabular}}}
\end{center}
\label{table_tc_ablation}
\end{table}

We mainly present the ablation study on Total Capture dataset for evaluating the efficiency of each contribution of our proposed strategy. To avoid the influence of 2D estimation error and better analyze our method, in the ablation study we use 2D GT pose sequence as input. The qualitative results are shown in Table \ref{table_tc_ablation}.

\noindent \textbf{On Sensor Fusion.}
\quad First, we analyze the influence of the $SF^{2}$ module and inertial energy in the optimization module. This ablation setting is the same as RTO. The results are in the second column of Table \ref{table_tc_ablation}. We can observe that compared to the final result of RTOF with $23.2mm$, the MPJPE of RTO is $40.6mm$. This demonstrates that sensor fusion in kinematic space is indispensable for reducing estimation error. Furthermore, it is also shown that the method with $SF^{2}$ gets a result of $34.5mm$. The performance is greatly improved compared with the baseline result of $53.7mm$. This reveals that the inertial information is efficiently leveraged for correcting the human joint locations.

Despite the improvement, the proposed sensor fusion has an important shortcoming that the single frame fusion introduces more temporal jitters in the output sequence. This reflects in the results in the third column of Table \ref{table_tc_ablation}. For the method with $SF^{2}$, although the MPJPE is reduced by a large margin, the MPJJE and MPJAE are increased compared to the baseline. The temporal jitters need to be further eliminated by exploiting correlations across adjacent frames in the refined 3D pose sequence.

\noindent \textbf{On Temporal Optimization.}
\quad This part analyzes the value of the proposed fragment-based temporal optimization for the refined 3D pose sequence by $SF^{2}$ module. We first compare the result of RTO with the baseline in Table \ref{table_tc_ablation}, especially on the metrics of motion acceleration and jitters. Next, we can also observe that the result with RTOF has a much larger reduction compared to the result with $SF^{2}$. The MPJJE is decreased by over $60\%$. Compared to the baseline result, the jitters are also reduced and the 3D HPE performance is improved by over $50\%$, strongly proving the value of our proposed framework.

\noindent \textbf{Qualitative Results.}
\quad We draw the estimated 3D human pose compared with the ground truth pose in Figure \ref{fig_pose}. The baseline results have significant errors on limbs due to the depth ambiguity when performing 2D-to-3D lifting. The proposed $SF^{2}$ module can correct these errors using IMU-measured limb orientations. Furthermore, the bone stretching issue is well solved via inverse and forward kinematic processes. The estimated skeletons are finally optimized in 3D space with hybrid energy functions. Thus, our RTOF framework achieves satisfactory 3D HPE results.

\noindent \textbf{On Real-Time 3D HPE.}
\quad As shown in Table \ref{table_N}, we investigate the impact of fragment length $N$ for online optimization. We report the number of output frames per second (FPS), MPJPE and MPJJE in the case of $N = 200, 100, 50, 20$ respectively. The results demonstrate that longer fragments need more time to infer. This is mainly because the optimized parameters increase. Another phenomenon is that the smaller $N$, the lower MPJPE. This shows the effectiveness of our fragment level optimization. However, the temporal jitters can be decreased when choosing a larger $N$. To balance the performance on accuracy and smoothness, we finally set $N=50$ in this work, also enabling real-time 3D HPE with negligible delay.

\begin{table}[!t]
\caption{Analysis on the selection for the length $N$ of the optimized fragments. FPS is the number of output frames per second. The inputs are all ground truth 2D poses.}
\begin{center}
\scalebox{0.25}{
\resizebox{\textwidth}{!}{
\begin{tabular}{c|ccc}
\hline
N & FPS & MPJPE & MPJJE \\
\hline
200 & 227 & 23.64 & 59.01\\
100 & 423 & 23.27 & 60.03\\
50 & 743 & 23.23 & 63.84\\
20 & 1256 & 23.19 & 77.31\\
\hline
\end{tabular}}}
\end{center}
\label{table_N}
\end{table}

\noindent \textbf{On Weights of Energy Functions.}
\quad As shown in Table \ref{table_kv}, we investigate the impact of the weights of energy functions in temporal optimization.  When $k_{V}$ equals 1, it means that the inertial term is not used and the motion jitter error is very large. This reflects the importance of the acceleration and jitter energy terms built from IMUs. When using both visual and inertial terms, $k_V=0.9$ achieves the best results on two metrics.

\begin{table}[!t]
\caption{Analysis on the selection for the weight $k_{V}$ of the visual energy. The inputs are all ground truth 2D poses.}
\begin{center}
\scalebox{0.42}{
\resizebox{\textwidth}{!}{
\begin{tabular}{c|ccccc}
\hline
$k_V$ & 1.0 & 0.9 & 0.8 & 0.7&0.6\\
\hline
MPJPE &23.29& 23.23 & 23.23 & 23.24&23.26\\
MPJJE &104.9& 63.84 & 68.46 & 72.74&77.19\\
\hline
\end{tabular}}}
\end{center}
\label{table_kv}
\end{table}

\section{Discussion}

Figure \ref{fig_curves} illustrates some error curves with the comparison of the baseline and the proposed RTOF framework. On the left, we draw the average joint error curves of two sequences. We can observe that the orange curve is all below the blue one. Also, most of the curve peaks are well decreased, which demonstrates the ability of our method for removing the outliers. This is because IMUs can provide relatively reliable bone local rotations to refine the incorrect one in the initial lifted 3D pose. This can further be proved by the curves on the right of Figure \ref{fig_curves}. We draw the rotation angle error curves of the human right knee in two motion sequences. The angle estimation performance is greatly improved. Moreover, the fluctuation of the orange curve is much smaller than that of the blue curve, which reflects that the output 3D human pose sequence of RTOF is more smooth than that of the baseline. The proposed framework can estimate physically more plausible motion. In applications such as medical rehabilitation assessment or fitness guidance, where smooth and accurate human motion parameters need to be measured, our method will be valuable and effective. There are also some limitations in the proposed framework. The IMUs need to be calibrated with the camera before motion capture. Also, the many-to-one backbone method introduces some delays in inference. In the future, we will continue to study a causal method to better avoid delays. Also, how to utilize the rule of dynamic to further improve the result is also worth exploring.

\section{Conclusion}

In this work, we present RTOF, an online temporal optimization and fusion for 3D human pose estimation from video and IMU data. The lifted 3D poses are first refined in each frame via a kinematic-based algorithm using IMU orientation information. The refined pose sequences are then optimized temporally to decrease jitters, which can smoothen the final output and make it plausible physically and biomechanically. Comprehensive experiments on two datasets demonstrate that our method achieves superior performance over most previous state-of-the-arts. Future work may focus on its applications such as rehabilitation assessment and action recognition.

\section*{Acknowledgments}
This work has been supported by the NSFC grants 62176156 and Deepwise Healthcare Joint Research Lab, Shanghai Jiao Tong University.

{\small
\bibliographystyle{model2-names}
\bibliography{refs}
}

\end{document}